# A framework for river connectivity classification using temporal image processing and attention based neural networks

Timothy James Becker[1], Derin Gezgin[1], Jun Yi He Wu[1], Mary Becker[2]
[1]Connecticut College, [2]Connecticut Department of Energy and Environmental Protection

**Abstract**
Measuring the connectivity of water in rivers and streams is essential for effective water resource management.  Increased extreme weather events associated with climate change can result in alterations to river and stream connectivity.  While traditional stream flow gauges are costly to deploy and limited to large river bodies, trail camera methods are a low-cost and easily deployed alternative to collect hourly data.  Image capturing, however requires stream ecologists to manually curate (select and label) tens of thousands of images per year.  To improve this workflow, we developed an automated instream trail camera image classification system consisting of three parts: (1) image processing, (2) image augmentation and (3) machine learning.  The image preprocessing consists of seven image quality filters, foliage-based luma variance reduction, resizing and bottom-center cropping.  Images are balanced using variable amount of generative augmentation using diffusion models and then passed to a machine learning classification model in labeled form.  By using the vision transformer architecture and temporal image enhancement in our framework, we are able to increase the 75% base accuracy to 90% for a new unseen site image.  We make use of a dataset captured and labeled by staff from the Connecticut Department of Energy and Environmental Protection between 2018-2020.  Our results indicate that a combination of temporal image processing and attention-based models are effective at classifying unseen river connectivity images.

## Introduction

The connection of flow is an important driver in river and stream ecosystem function [Hynes 1975, Vannote et al 1980].  Maintaining natural patterns of river connectivity is essential to maintaining healthy populations of riverine species and is critical to the survival of certain species [Bunn & Arthington 2002, Wohl, 2017].  Alteration to natural river connectivity patterns occur from man-made structures, such as dams, culverts and groundwater withdrawals, as well as from extreme weather events due to climate change.  These alterations result in stream habitat alteration and threaten stream biodiversity [Carlisle et al 2011].  Effective water resource management requires monitoring river and stream connectivity conditions.

Traditional methods using stream gaging systems to measure stream flow and connectivity in the United States (U.S.) are expensive to deploy and cover less than 1% of rivers and streams across the U.S. [USGS, 2023].  They are typically deployed on large perennial streams providing limited information on small headwater streams [Zimmer et al 2020, USGS 2025].  Small headwater streams contribute approximately 70% of the mean-annual water volume in the U.S. [Alexander et al 2007].  Headwater streams also contribute a profound influence on the water quality and biological conditions in downstream waters but are particularly vulnerable to disruption [Meyer et al 2007].  In addition, traditional stream flow gages do not adequately describe certain connectivity conditions that can have different consequences for ecological processes [Zimmer et al 2020].  For example, disconnected streamflow resulting in zero-flow readings at gages can range from completely dry conditions leading to the extirpation of certain sensitive riverine species to disconnected pools and riffles which can provide refugia for riverine species facilitating recovery after surface flow resumes [Stubbington et al 2017, Malish et al 2023].  With widespread changes in disconnected events being observed, particularly in small headwater streams [Zipper et al 2021] as a result of climate change, it is imperative that additional methods augment stream gaging systems to help better inform water resource managers on river connectivity.

Hourly images of streams captured with trail camera is one method being used to address the challenges of more broadly monitoring stream flow and connectivity conditions across the U.S. [Bellucci et al 2020, Gupta et al 2022].  We leverage an image dataset collected in Connecticut, USA [Bellucci et al. 2020].  Images were reviewed by stream ecologists and labeled into six categories that describe river connectivity conditions related to ecosystem function ranging from varying disconnected to connected conditions (Figure 1).  This method has been successfully used to inform and prioritize management actions to restore river and stream connectivity, however manually labeling the images is labor intensive and fatiguing for staff scientists [Bellucci et al., 2020].

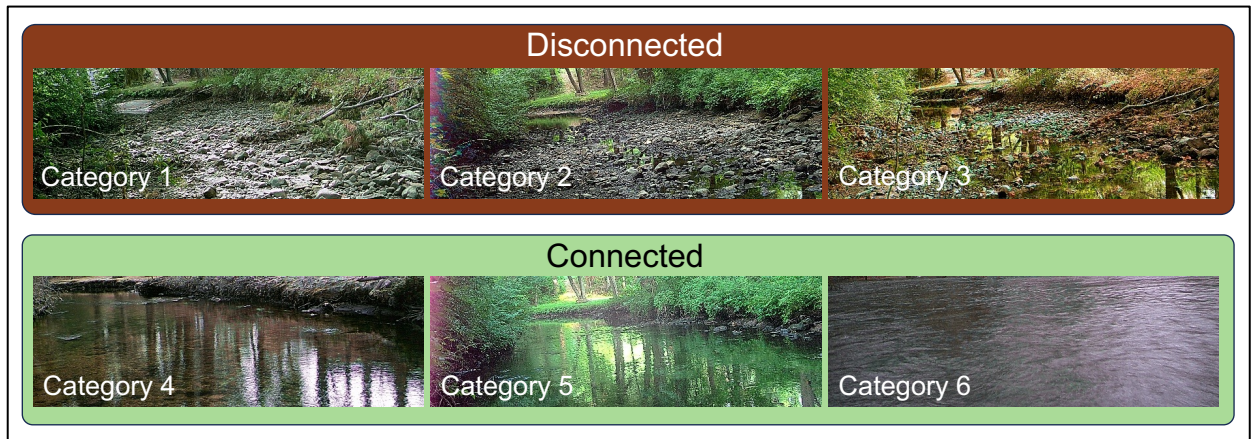

**Figure 1:** Disconnected state is shown in the top brown panel where we have Category 1, Category 2 and Category 3. Below the Connected state shown in green illustrates Category 4, Category 5 and Category 6 examples. All categories here are taken from the same site but at different time points (and deployments) showing the total span of all six categories across time.

The related problem of stream flow regression has shown progress using Deep Convolutional Neural Net architecture (DCNN) on color images [Gupta et al. 2020]. In their work Gupta et al. require images for every site and use a stratified data partitioning scheme that separates temporal flow states for each site. Other methods also published for this related problem included pixel-based approaches with in-frame scales [Noto et al. 2021]. In both of these works, a machine learning model is constructed for a single site using supervision in the form of additional data from a flow gauge (or the in-frame scale). These methods are both considered data driven models according to the definitions outlined in the guide written by Gauch and Lin [Gauch and Lin, 2020]. In our work, we formulate the general river connectivity classification problem which is related to streamflow but differs in its goal, inputs and management use cases [Bellucci et al. 2020]. Our problem is not specific to a certain site, requiring performance to be estimated on a completely unseen site image. Unlike the streamflow regression problem, the general river connectivity classification problem seeks to classify the categories listed in figure 1 which are numbered 1-6 or in binary form the groups disconnected and connected.

Image quality has a huge impact on the ability to classify an image but must be narrowly defined in order to create automated metrics. We focus our work to include only images that meet all criteria show in figure 2 which eliminates blurred, flared, extremely exposed, triggered or erroneous time-stamped images. Even after the low-quality images were filtered and bottom center auto-cropped, we found the quality and quantity of complex foliage-based shadow and highlight regions in each image made determining the water connectivity difficult. We looked to image enhancement literature from real-time video to decrease the effects of foliage and shadow in our images and developed an innovative method that exploits the temporal nature of the input images[Wu, et al. 2020].

In addition to image filtering and enhancement, we also made use of recent advances in machine learning image classification by implementing both DCNN and vision transformer models (ViT) that make use of attention mechanisms between image patches [Krizhevsky et al, 2017, Vanswani et al. 2017]. In each of these modeling methods, lower-level features such as foliage, rocks, water surfaces are implicitly learned and do not require domain-based modeling such as discussed by Gauch and Lin [Gauch and Lin, 2020]. We found that the newer ViT architecture while having increased training time, had better estimated accuracy using full hold out site image dataset.

In this paper, we detail methodology that uses traditional image processing along with image enhancement, recent machine learning image classification architectures and generative image augmentation to automate the image based general river connectivity problem. We show that our method is effective by utilizing a multi-year labeled dataset from the Connecticut Department of Energy and Environmental Protect from the years 2018-2020. We conclude with a discussion of the accuracy of the method as well as modeling generalizability and future directions.

## 2A Methods: Problem, Partitioning and Overview

**Problem Statement**
We define the river connection problem by first detailing the input images. Each image is expressed as a numeric matrix of the form: $X \in \mathbb{R}^{h \times w \times c}$: pixel height $h$, pixel width $w$ and pixel color channel(s) $c$. We let $\left[X_{s_1}^{i1}, \dots, X_{s_1}^{n_1}\right]$ be the timeseries of $n_1$ images taken from site $s_1$, while in general we will have a total of $m$ unique sites which represent a specific camera perspective of a particular space. Because we wish to train models that can learn lower-level features and classify new unseen sites, we must partition our total $m$ sites so that our final estimate does not share any sites used in training (or the intermediary testing phase). Our total dataset then is expressed as the set of sites: $\{[X_{s_1}^{i1}, \dots, X_{s_1}^{n_1}], \dots, [X_{s_m}^{i1}, \dots, X_{s_m}^{n_m}]\}$. Note that some sites will have many images, while others which may represent newer deployments (or deployments that are no longer being conducted) will have very few. Because this is a supervised problem formulation, we also have the corresponding labels for each image as the set of vectors: $\{[Y_{s_1}^{i1}, \dots, Y_{s_1}^{n_1}], \dots, [Y_{s_m}^{i1}, \dots, Y_{s_m}^{n_m}]\}$, where each value is taken from the label set: $L = \{1,2,3,4,5,6\}$. Because camera deployments are at a specific time and spatial position, observations of different connection states based on environmental and temperature conditions will naturally create a complex discrete multinomial distribution which we can define as the random variable $\mathbb{Y}$:

$$P(\mathbb{Y} = y)_{y \in L} = \left[\frac{\sum_{i=1}^{m}\sum_{j=1}^{n_i} b(Y_{s_i}^{n_j}, y)}{\sum_{i=1}^{m}\sum_{j=1}^{n_i} 1}\right] : b(y, l) = \begin{cases} 0, & y = l \\ 1, & else \end{cases} \quad (1)$$

We denote the set of site ids as: $S = \{s_1, \dots, s_m\}$. We seek to train a model which will learn a non-linear mapping of the input images to the corresponding label or to infer $Y$ from $X$.

**Partitioning**
To measure the accuracy of the mapping, we must partition our data into three subsets of sites such that each subset has a discrete multinomial distribution as close to the total distribution detailed in equation (1). We count the number of labels from any given site $s_i$ as:

$$H(s_i, l) = \sum_{j=1}^{n_i} b\left(Y_{s_i}^{n_j}, l\right) \quad (2)$$

We can then measure any subset distribution divergence as the distance function between two pairs using either the absolute difference of the pooled labels from each subset of sites or we use the square root of the squared difference:

$$\delta(u, t) = \left|\sum_{l=1}^{|L|} \left(\sum_{i \in u} H(S_i, l) - \sum_{j \in t} H(S_j, l)\right)\right| \quad (3)$$

$$\delta_2(u, t) = \sqrt{\sum_{l=1}^{|L|} \left(\sum_{i \in u} H(S_i, l) - \sum_{j \in t} H(S_j, l)\right)^2} \quad (4)$$

We then define two partition sizes split parameter $\theta$ and total images as $\gamma$
$$p_u = \lfloor \theta \gamma \rfloor, p_t = \lfloor (1 - \theta)\gamma \rfloor \quad (5)$$

We then find the minimally disturbed site id subsets for partition sizes $p_u, \gamma$ by randomly selecting site ids of size $p_a$ and then keeping track of the minimal distance from equation (3) or (4), which is an approximation of:

$$\underset{\forall u, t \subset S, |u| + |t| = |S|}{\text{argmin}} \{\delta(u, S) + \delta(t, S)\} \quad (6)$$

**Overview**
To solve the general classification problem, we make use of several steps to ensure high quality data is entering the machine learning framework. We first load data by using the metadata and file name to keep track of the image number and its association with a specific site id and time stamp. In this way we have a spatial and temporal dataset shown in the left of figure 2. The partitioning step occurs after filtration and image enhancement since some of the images will be lost due to quality control and we want to compute our total image set on image that pass all filtration. We look for images that are (1) too bright (over exposed) (2) too dark (under exposed), (3) lacking color information (black and white taken from infrared camera modes), (4) blurred (usually from weather and temperature based condensation), (5) flared (due to the angle to sun which is can be hard to predict during deployment with additional setup and compass usage), (6) triggered (which are additional images that are usually people or animals walking past the camera) and finally images where the camera has not been set correctly before deployment which means the image ordering and time are not reliable. The remaining pictures are then enhanced and bottom center auto-cropped before given to the machine learning framework shown in figure 2. We found that the passed images still contained incredibly complex shadows and highlights from foliage, tree trunks, branches and sun reflections on water surfaces. We looked to recent literature in real-time object tracking that used interframe images to remove shadows from the target object (people) [Wu et al. 2020]. We generalizable the removal to become a localized gradient to lower luma variance which has the effect of decreasing local shadows and highlights while adding overall detail.

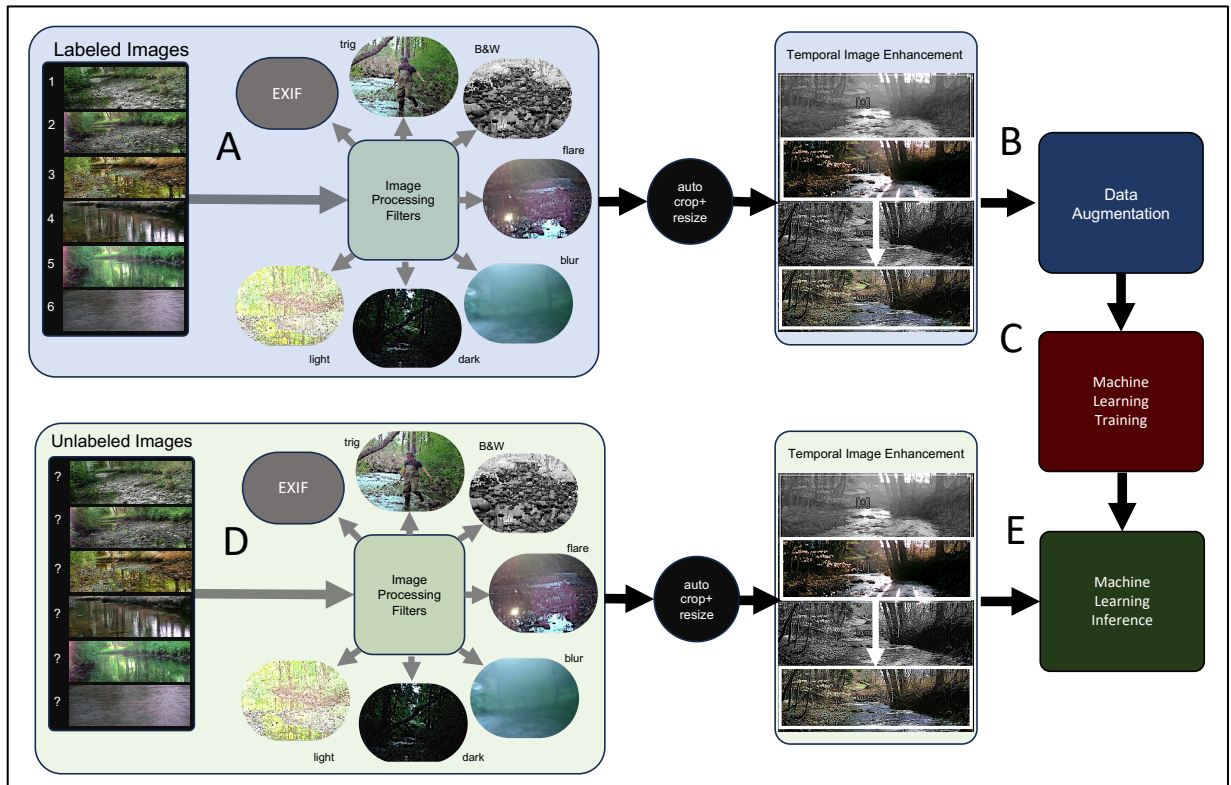

**Figure 2: (A)** The training phase filters low quality images before bottom-center auto-cropping and image enhancement is applied. **(B)** The data augmentation framework can generate more images for labels that are lower in frequency such as 1 or 6. **(C)** The augmented data is then given to the training model to fit and then saved for future inference. **(D)** The same preprocessing is used as in (A) but now the framework will not have labels. Instead, the model file is loaded and then proceeds to directly to inference in **(E)**.

## 2B Methods: Image Enhancement

To increase the image quality for classification, we use the exchangeable image file format (EXIF) time stamps to match and cluster the hourly images within an aggregation window *a* and then use a variation of the shadow elimination algorithm detailed by Wu et al. [Wu, et al. 2020]. In their work Wu et al. are seeking to completely eliminate subject shadows in real time video by looking at interframe differences (historically windowed). Because real-time usage is not required in our application, we instead take $\frac{\alpha}{2}$ images *before* and *after* each image we seek to enhance. This provides a controllable mixture as opposed to complete removal. By splitting into the *YCrCb* color space we then can average the luma channels of the $\alpha$ images (*Y* from the *YCrCb* color space) before mixing back into the original by a controllable amount. In effect this lowers the total variance of the luma channel (*Y*) in areas of prominent shadow and highlight. The temporal enhancement effect can be seen in figure 3 below.

We take for each input image of the form: $X \in \mathbb{R}^{h \times w \times c}$ which normally has the color channels set to (blue, green, red) in the OpenCV implementation we use. These color channels are then converted to *YCrCb* color space and the *Y* or luma channel is enhanced across the $\alpha$ images. We denote $X_{s_i}[luma] \in R^{h \times w}$ as the luma channel images for site $s_i$. We then find the average luma across the $\alpha$ images and then mix the difference to each image using parameter $\beta$:

$$\mu_{s_i}[luma] = \left(\frac{1}{\alpha}\right)\left(\sum_{j \in \alpha} X_{s_i}^{n_j}[luma]\right) \quad (7)$$

$$X_{s_i}^{n_j}[luma]_{j \in \alpha} = X_{s_i}^{n_j}[luma] + \beta\left(X_{s_i}^{n_j}[luma] - \mu_{s_i}[luma]\right) \quad (8)$$

The enhanced luma images are then converted back to the native color space for further processing.

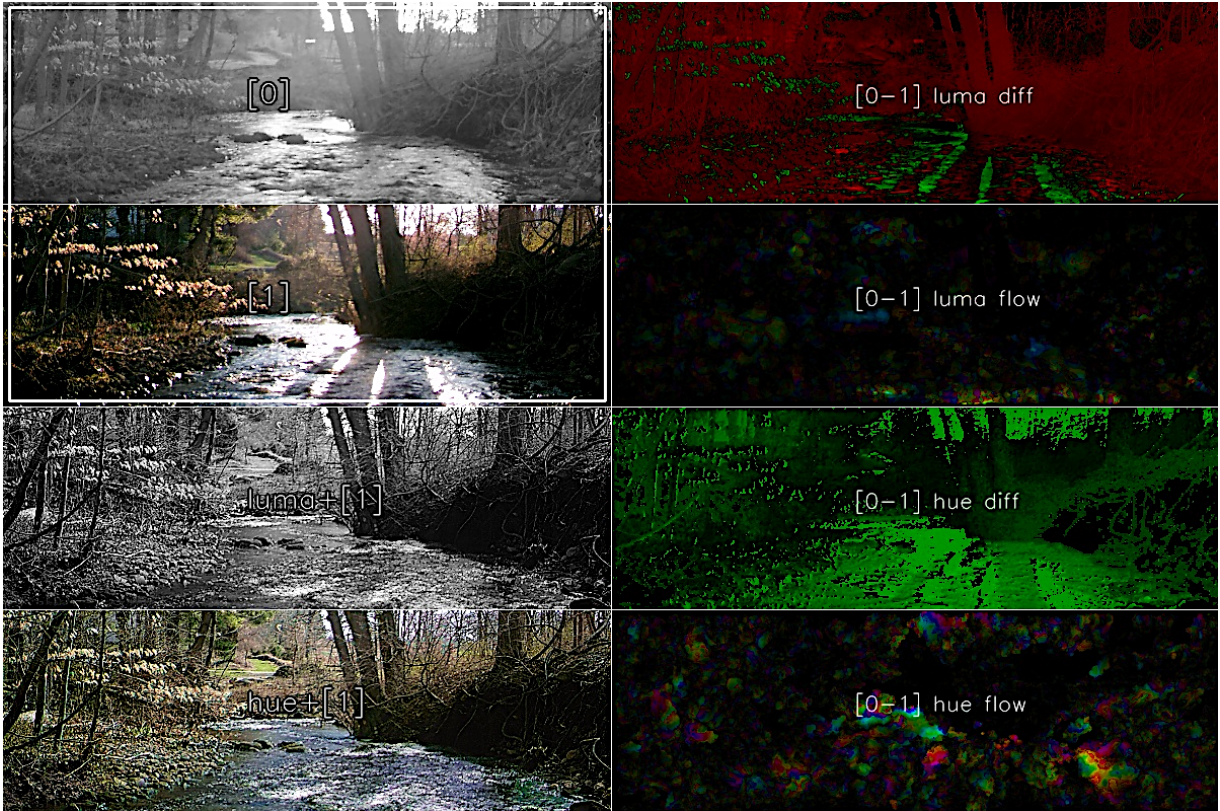

**Figure 3:** [0] shows the temporal image before the currently processed image [1]. Image [0] has very low color variance and almost appears black and white in comparison to image [1] but it contains more details in the riverbank region to the right as well as more details in the water surface. These details take the form of luma gradients and by looking at differences between images (shown on the right as [0-1] luma diff, and the optical flow we can determine the major changes in the sun as passing through foliage or in this case the tree trunk in the foreground. When we make use of image [0] to enhance image [1] as shown in luma+[1] and hue+[1] we can see that shadow and highlights are reduced and detail is increased in the riverbanks and throughout the entire image.

## 2C Methods: Machine Learning Framework

We implemented and then tuned two well established methods for classification: deep residual convolutional neural networks (RESNET) and vision transformers (ViT). Both methods have seen many successful uses in image processing classification tasks, while neither has been deployed in the general river connectivity classification task [Dosovitskiy et al., 2021, Khan et al., 2021, Gupta et al., 2022]. Overall, we found that both methods are suitable for learning the water connectivity patterns for our problem but have different model sizes which determines the practicality of the inference.

**Deep Residual Convolutional Neural Networks**
DCNNs have long been used for computer vision classification tasks [He et al., 2016]. The core methodology involves structuring convolutional layers using a kernel (or window around each pixel) to extract the important gradients from the image set. Convolutional layers essentially are kernel filters that have unfound coefficients and the training process under supervision will train competing filters that will converge on important visual gradients which are used to partition the space of instances across the label classes boundaries. We employed a RESNET style model for our work which adds residual connections (similar to attention connections) to a regular DCNN which has been shown to generalize well on image data [He et al., 2016].

**Vision Transformers**
ViTs have emerged as a powerful deep-learning architecture for computer vision tasks which leverage self-attention [Vaswani et al., 2017]. Unlike traditional convolutional neural networks, vision transformers seek to establish the relationship between regions of the input in a given layer by dividing the space into patches and processing them as sequences of tokens. For certain applications this architecture was shown to match or exceed the performance of DCNNs but with faster training time [Dosovitskiy et al., 2021]. Kan et al. detail many variations on the initial vision transformers and show a wide application across many image classification tasks [Khan et al., 2021]. Although the related work from Gupta et al. made use of DCNN they did not attempt to employ ViT. We hypothesized ViT would provide superior performance to DCNN on our data because of the self-attention across patches which could better generalize patterns in the images like foliage, sky, river bed rocks, stones, water reflection, etc.

**Training Procedures**
For our work we focused on a very small set of hyperparameters and sought to build models that would be small in size and deployable on general purpose computing [Keras, 2025]. Our final model training results which are detailed in **section 3**, show that reasonably sized models can be built and trained within a few minutes on a standard Linux workstation with an Nvidia A4000 GPU or better. The trained models are small enough in size that they can generate inference using a standard workstation or laptop that has sufficient RAM (>16GB) and without a GPU. We envision that with more work in this area we could achieve CPU-only inference with very high accuracy >90% between connected and disconnected streams and rivers.

## 2D Methods: Data Augmentation

**Imbalances in Labeled Data**
An imbalanced dataset can cause an image classifier to underperform on minority classes while achieving misleadingly high accuracy on majority classes [Leevy et al., 2018]. This makes the model unsuitable for our goal of monitoring river and stream connectivity to inform water resource managers. Outliers, such as labels 1 and 6, are critical indicators of potential issues in water resource management. Therefore, reshaping the data distribution is essential to ensure an unbiased model.

To address imbalanced datasets, researchers commonly use data augmentation. For image classification models, oversampling and geometric transformations—such as flipping, color adjustments, cropping, rotations, translations, noise injection, and image mixing—are typical approaches [Shorten & Khoshgoftaar, 2019]. These techniques introduce new challenges to models by encoding invariances absent in the original training set, reducing the gap between the training and validation sets. A notable application of augmentation is found in "ImageNet Classification with Deep Convolutional Neural Networks." Krizhevsky et al. used augmentation to expand the training set by more than 2000 times through random 224 × 224 crops, horizontal flips, and RGB intensity transformations [Krizhevsky et al., 2012].

**Generative Image Augmentation**
While traditional hand-crafted data augmentation can improve performance, it often fails to capture the dataset's contextual distribution, as augmented images only locally differ from the original image in a small area [Li et al., 2023]. This is especially true when augmentation is based on a single image, resulting in images that are either too dissimilar (leading to performance degradation) or too similar (causing overfitting). In that case, the hand-crafted augmentation becomes a restriction [Li et al., 2023].

Denoising Diffusion Models (DDM), introduced by Ho et al. are neural networks that use a parameterized Markov chain trained via variational inference to generate samples that match the data distribution after a finite number of steps [Ho et al., 2020]. The model features a forward diffusion process, which adds noise, and a reverse diffusion process, which denoises it [Ho et al., 2020]. Unlike general adversarial models (GANs), DMs offer advantages such as distribution coverage, a stable training objective, and scalability, though at a higher computational cost [Dhariwal & Nichol, 2021]. Recent work by Dhariwal and Nichol has closed performance gaps, achieving state-of-the-art scores on ImageNet and LSUN while maintaining better dataset coverage [Dhariwal & Nichol, 2021].

For DM based augmentation, Azizi et al. demonstrated that augmenting ImageNet with diffusion models improved classification accuracy. By fine-tuning an Imagen model, they achieved state-of-the-art FID (1.76) and Inception Score (239) at 256x256 resolution, and new state-of-the-art classification accuracy scores with only synthetic data: 64.96% at 256x256 and 69.24% at 1024x1024. Synthetic data improved ResNet-50 performance by 1.78%, from 76.39% (real data only) to 78.17% (synthetic + real), comparable to Krizhevsky et al.'s augmentation boost. Li et al. tackled a similar problem but addressed image diversity and semantic consistency using semantic-guided generative image augmentation with DMs (SGID) [Li et al., 2023].

**Generative Image Models for Streams and Rivers**
We used the existing literature on GANs and DDM to experiment with our own DDM based generative augmentation model. To properly feed the augmentation model, we created an augmentation pipeline that used traditional geometric methods such as rotations (5° to 30° in 5° increments) and horizontal flips. Initial observations showed that 5° provided adequate variation while limiting rotations to 30° prevented excessive distortion of the original dataset. In our data care had to be taken to limit rotation since vertical flips would create upside-down rivers. In our case, 5° rotations and horizontal flips were used because they reflect the variability in trail camera placement and river orientations. Trail cameras are tethered to a tree, "high enough" from a "flood zone" [Bellucci et al., 2020], sometimes with a slight tilt, while horizontal flips allow diversification since rivers come in all directions. Our augmentation pipeline involves the following steps: (1) pad the image, (2) randomly rotate by 5° (within limited range), (3) resize (1.3x per rotation), (4) center crop, and (5) horizontal flip (if not already applied).

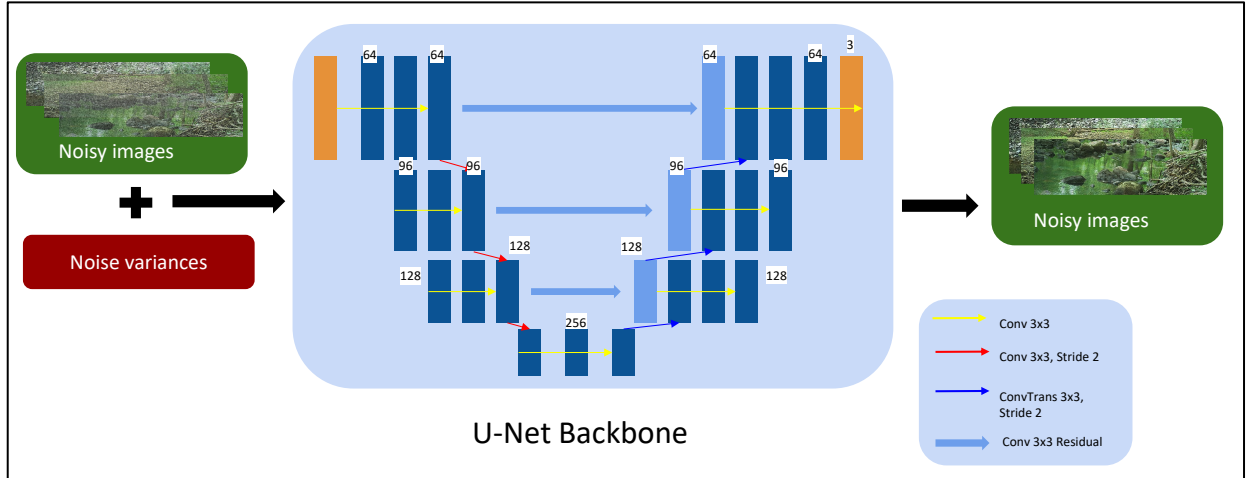

**Figure 4:** Generative augmentation pipeline for synthetic image generation. The input images are derived from real images that undergoes traditional augmentation to present the underlying visual information for the diffusion process to sample from. Diffusion models start with total noise and attempt to learn the patterns of denoising which will result in a diffusion image that has some of the visual properties of the input images but will contain many unique variations and not look exactly like any one image from the input set.

## 3 Results

**Performance Metrics**
An important part of developing any machine learning system is to define one or more functional metrics. We looked at *F1* accuracy and *F1* disconnected accuracy which we found to be more challenging than total accuracy. The *F1* disconnected accuracy in particular provides a lower bound of the model to classify a new unseen image as being disconnected. Since the disconnected state is observed less frequently in the deployment of these trail cameras, we took care to balance inputs to the models to create an unbiased estimate of performance.

We define the class $A$ label precision and recall and $F1$, given a prediction vector $\bar{Y}$ and true vector $Y$ as:

$$prec_A = \frac{|\bar{Y}=Y=a|}{|Y=a|}, rec_A = \frac{|\bar{Y}=Y=a|}{|\bar{Y}=a|}, F1_A = \frac{2(prec_A * rec_A)}{(prec_A + rec_A)} \tag{9}$$

Where $|\bar{Y} = Y = a|$ is the number of correct predictions for class $A$, $|\bar{Y} = a|$ is the number of predicted $A$ labels and $|\bar{Y} = a|$ is the true number of $A$ labels.

We define the combined $m$ class label F1 accuracy, given a prediction vector $\bar{Y}$ and true vector $Y$ as:

$$F1 = \frac{m \prod_{i=1}^{m} F1_i}{\sum_{i=1}^{m} F1_i} \tag{10}$$

**Limitations of Water Connection Classes**
To support our effort on the 2-class problem, we first explored the limitation of six of the class labels since the class 3 and 4 were very difficult for human classifiers to discriminate. We wanted to verify that this same pattern was present in our machine learning models and to also help to explain any performance that was lost in our 2-class model [Bellucci et al. 2020]. For this experiment we used preprocessed raw data (no augmentation) without temporal enhancement (0 hour) and used hyperparameter search on a RESNET model architecture only since in general the model types behave in a similar manner towards both basic augmentation as well as temporal enhancement. We then looked at the confusion matrices below in figure 5 for the resulting 6-class model to determine which class labels were most difficult for the machine learning model. We found that the model could find the 1 very easily but 2 and 3 were often confused while label 4 and 5 were easier than 6. We confirm that a 2-class problem (disconnected versus connected) is a reachable goal and the 6-class variant is considerably harder.

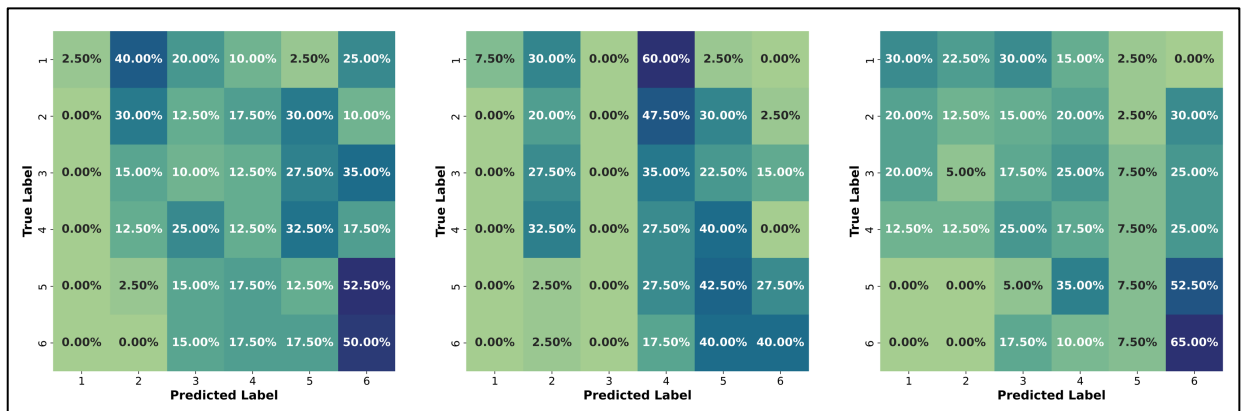

**Figure 5:** Full validation holdout confusion matrix for three ViT training runs of a six-class classification model. Attempting to apply classification to the classes directly is more challenging than the binary disconnected to connected classifier we have detailed. Early experiments show that labels are confused near each other but also the confusion shows a general bifurcation or split near 3-4 which is between the connected and disconnect labels number 1,2,3 against 4,5,6. Contrasted with **figure 6F** it is clear that a high performing connected-disconnected model can be built while the six-class version remains an open problem that our methods do not adequately address.

**Temporal Enhancement Effect**
To determine the effect of the temporal enhancement in **section 2B**, we devised a series of multiple experiments where we looked at 0 hours temporal enhancement (raw), 2 hours of temporal enhancement and 4 hours of temporal enhancement for unaugment data in order to isolate its potential effect on accuracy. The results are shown below in figure 6 where the temporal enhancement increases accuracy of classification regardless of the model architecture. In each experiment we used the partitioning strategy from **section 2A** and measured performance using only fully reserved images from the validation partition of site ids. This ensured each model had never seen any images from that site id set.

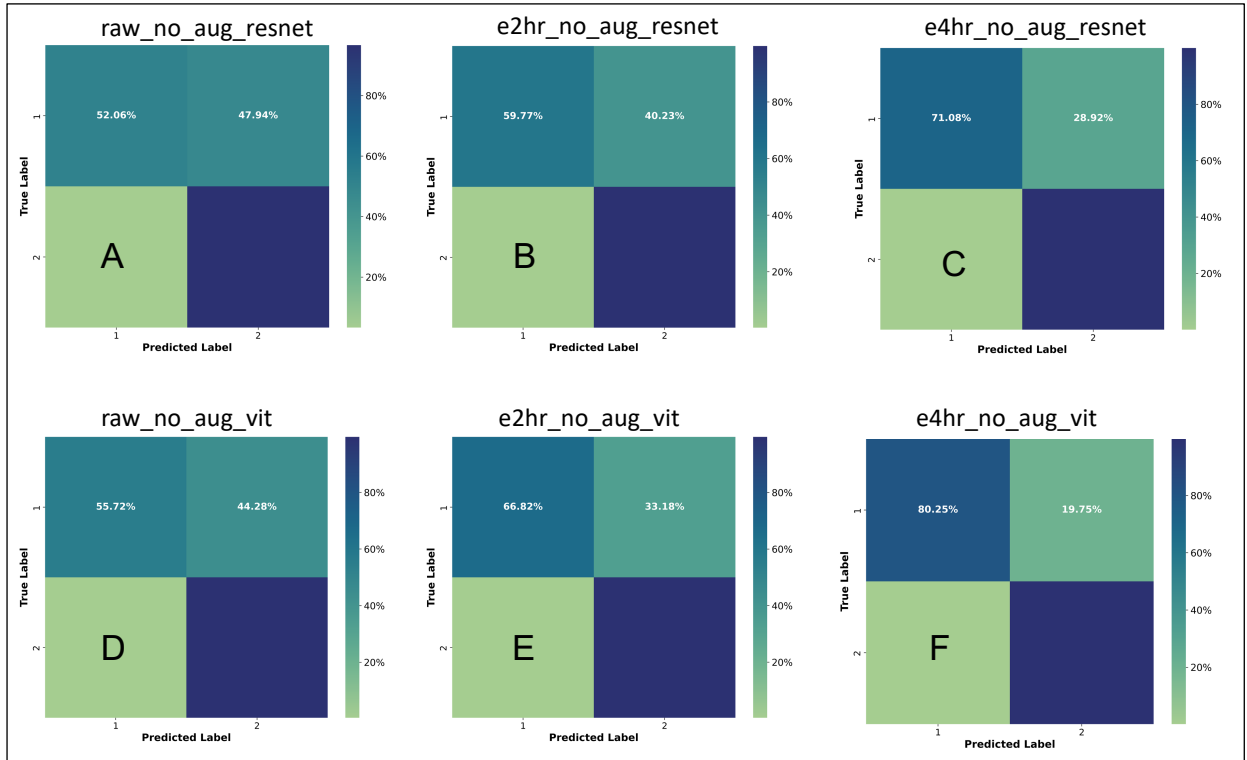

**Figure 6:** Comparison of RESNET DCNN and ViT models built with no augmentation in the training and test partitions and calculated using the full holdout validation dataset. **(A) raw_no_aug_resnet** is a DCNN with no temporal enhancement and no augmentation, **(B) e2hr_no_aug_resnet** is a DCNN with a 2 hour enhancement window applied and no augmentation, **(C) e4hr_no_aug_resnet** is a DCNN with a 4 hour enhancement applied and no augmentation, **(D) raw_no_aug_vit** is a ViT with no temporal enhancement and no augmentation, **(E) e2hr_no_aug_vit** is a ViT with a 2 hour enhancement applied and no augmentation, **(F) e4hr_no_aug_vit** is a ViT with a 4 hour enhancement applied and no augmentation. In both model architectures the temporal enhancement makes a significant contribution to accuracy in the disconnected class 1. This indicates that temporal enhancement provides more visual information for detecting disconnected stream and river images.

**Basic Augmentation Effects**

To determine the effect of basic augmentation (flipping and histogram equalization) versus raw training and test data (validation data is never augmented) we compared RESNET and ViT models built with and without augmentation as shown below in figure 7. Augmentation is conducted by measuring the imbalance in the individual class labels and then producing a multiple of the training and test set data partitions, followed by down sampling to ensure the final number of images (both original and augmented) achieve a similar duplicity. We found that basic augmentation provides lower overall accuracy, lower disconnected *F1* accuracy and lower combined connected and disconnected *F1* accuracy. Basic augmentation does not score higher than the raw data which could be a result of the histogram equalization which is not effective at providing new information but lowers total image variance, whereas the temporal enhancement across images increase performance because of its local application using hourly image differences controlled in the *YCrCb* color space.

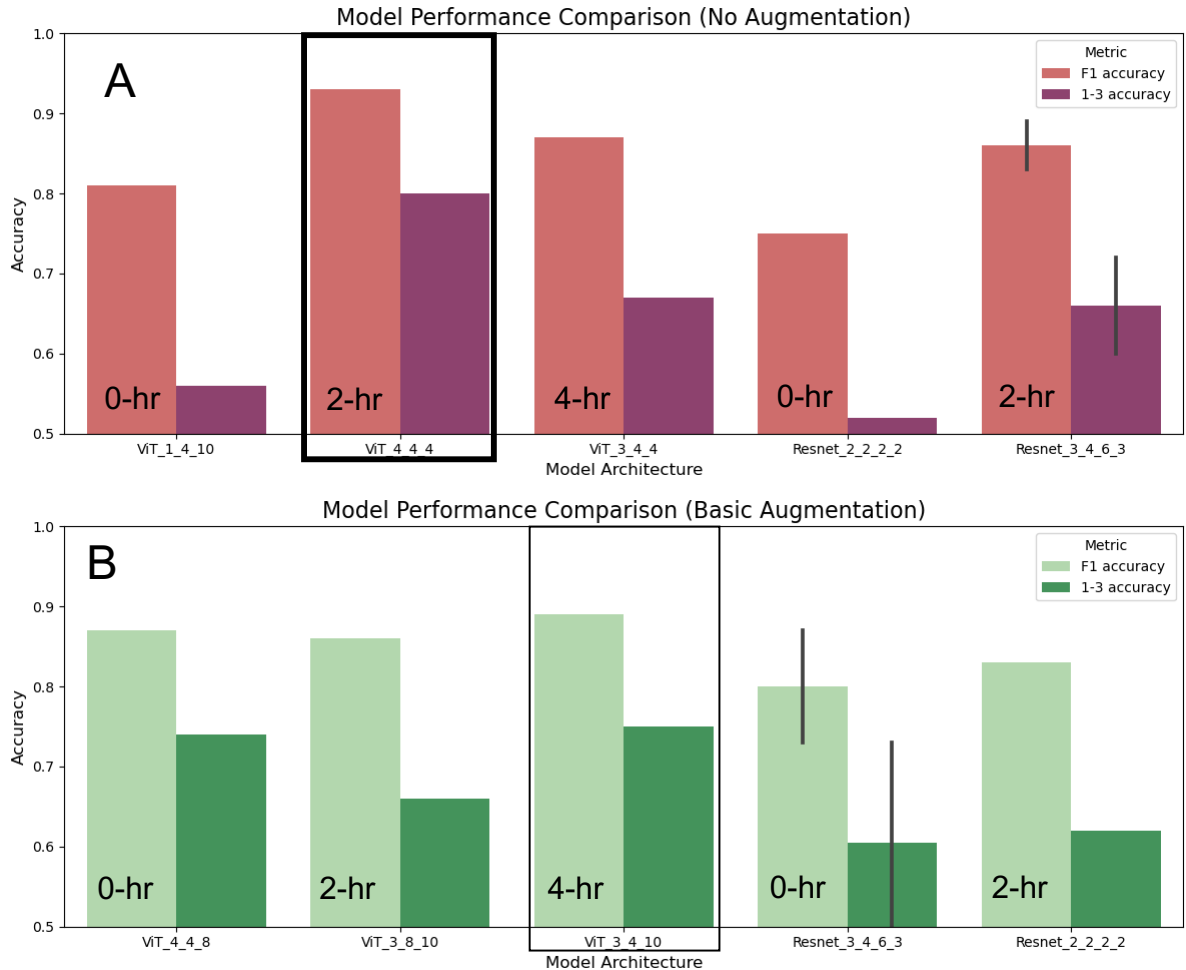

**Figure 7:** (A) Upper panel where no augmentation is used for the ViT and RESNET DCNN models in red. Temporal enhancement of 2 hours (shown with the black box) provides the highest disconnected F1 accuracy (lowest purple bar), as well as combined F1 accuracy and overall accuracy. (B) Lower panel where basic augmentation is used along with temporal enhancement of the RESNET DCNN and ViT models. Here the 4-hour temporal enhancement (shown with the black box) is the top performing model in our overall accuracy, disconnected F1 accuracy and combined F1 accuracy. Basic augmentation does not score higher than the raw data which could be a result of the histogram equalization which is not effective at providing new information but lowers total image variance, whereas the temporal enhancement across images increase performance because of its local application using hour image differences controlled in the $YCrCb$ color space.

**Generative Augmentation Effects**

Because the generative models produce new unseen images but draws them from a similar distribution as the training images in theory the augmentation should work better than histogram equalization which would lower the overall variance and image information across a set of images. But generating new images using diffusion models has not been previously in our challenging river and stream domain and so we had to explore modifications and several architectures in order to generate images that do not lower validation performance given that is the ultimate goal. In our work we partitioned our dataset by its label and then trained DDMs on augmented versions described in **section 2D**. The results were then compared against some of the input images to look for details like foliage, rocks, mud, water surfaces and reflections and overall composition. We found that with sufficient training we could effectively generate low frequency models with adequate visual quality as shown below in figure 8. Importantly, these new generated images provide the needed site-based diversification (which goes beyond linear geometric transforms) that a general river and stream connectivity system would need to learn low frequency visual characteristics in general.

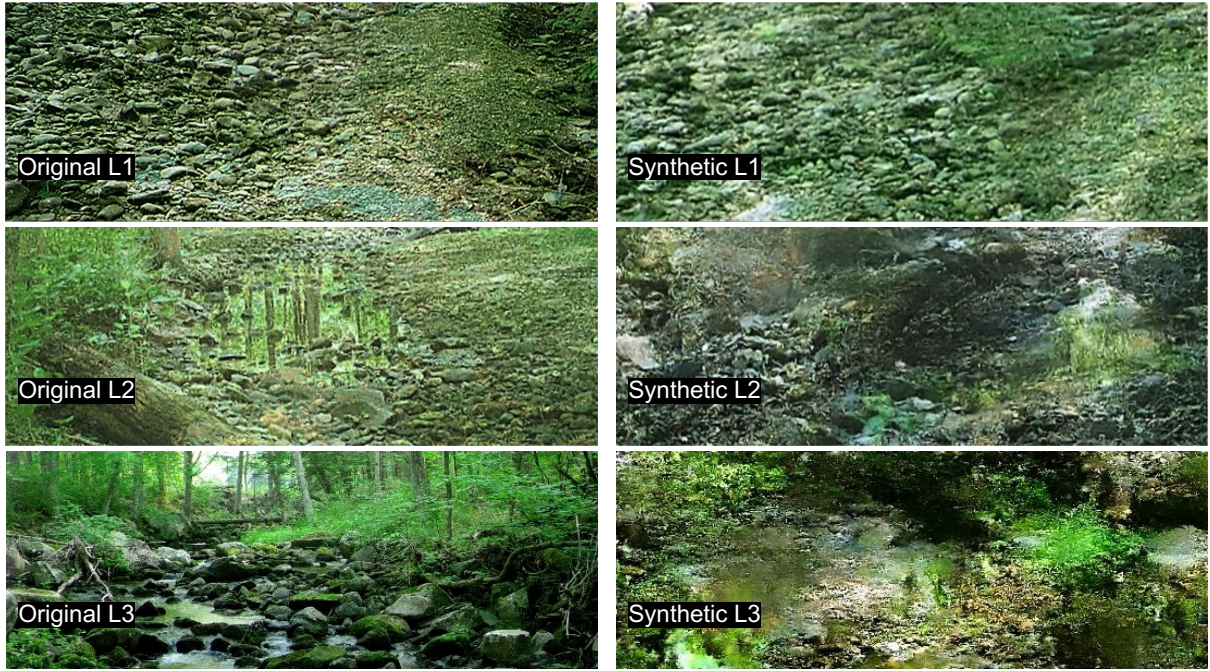

**Figure 8:** (Original L1) in the upper left is an image taken from a real observed label 1 image which has no connectivity. (Synthetic L1) in the upper right is an image generated by the DDM L1 model. (Original L2) middle left is an image taken from a real observed label 2 disconnected state. (Synthetic L2) middle right is an image generated by the DDM L2 model. (Original L3) lower left is an image of a real observed label 3 disconnected category. (Synthetic L3) lower right is a generative augmentation-based label 3 image from our DDM L3 model.

Late in our project we completed the DDM and generated three labels specific versions which we then integrated into our real dataset for balancing the class labels in the training and test partitions. We did not have time to then complete the full objective comparison and shown in figure 7. Instead, we trained a ViT with the medium level of temporal enhancement (2hr) which resulted in 0.73 *F1* accuracy which was lower than our ViT with medium level of temporal enhancement (2hr) which scored 0.87 *F1* accuracy. It is possible that better model training parameters would score higher but we did not have time to fully explore this area.

## 4 Discussion

We set out by describing and formulating the image based general stream connection problem. We provided a robust method to partition the dataset so that prediction can be measured in an unbiased manner on unseen sites as opposed to fitting a model to the data and then using the accuracy of the fit. The seven-filter preprocessing pipeline we developed effectively eliminated thousands of low-quality images from the CT DEEP 2018-2020 dataset, while our temporal image enhancement decreased local shadow and highlights and gave greater detail. By making use of newer machine learning techniques like vision transformers, we saw an in increase in accuracy for the system along with faster training time and smaller models. These efficiency improvements indicate that the ViT architecture is superior for this application.

Interestingly we did not see a clear improvement by the use of augmentation for this data set. It may be that for the 2-class problem of determining disconnect versus connected that our dataset contains enough images and with sufficient diversity to not need augmentation techniques. Our experiments with DDMs did not increase overall performance either, but this portion of our work was incomplete as we lacked the time needed to effectively search for ViT parameters on the larger dataset (that was created as a result of the augmentation).

When we explore the possibility of using all connection classes for a 6-class model, we determined confusion will naturally arise for labels 1-2, 3-4 and 5-6. A natural extension to our work is to build models for classes 1-2, 3-4 and 5-6 (3-class model) which is supported in evidence from the figure 5 confusion matrix. Additionally, it may be possible that the automated feature extraction offered by the DCNN and ViT although useful object recognition tasks, may be lower in a 6-class performance than an ensemble model that uses separate domain-focus classifications such as region of interest (ROI) texture classifiers that could estimate masks across an input effectively segmenting the image space into water, sky, rock, ect. There is future work in exploring this alternate feature engineering strategy. With lower-level features available we would also be able to explore alternate cropping strategies that could potentially derive more augmentation for less frequent labels and could work synergistically with our DDM models [Redmon et al., 2015].

The last and most directly applicable extension would be to add the ability to condition preexisting models with new input data in a semi-supervised framework. In this capacity, the information and datasets generated by CT DEEP and others could be leveraged in new locations where the topology, coloration, light or other foliage patterns diverge significantly such as artic or desert conditions. Given the simplicity and low cost of this data collection method, it seems likely that more training data will be made available in the near future which in turn will lead to better and more robust models using our framework. Because environmental habitats present a wide diversity and high visual complexity in comparison with simple human-centric objects, a mixture of traditional image processing techniques along with state-of-the-art neural networks is a valid means for building models targeting accuracy measures approaching human trained field staff.